\documentclass[journal]{IEEEtran}
\usepackage{graphicx,times,amsmath,amssymb,subfigure,lscape}
\allowdisplaybreaks 

\begin{document}

\title{Recommendations on Designing Practical\\ Interval Type-2 Fuzzy Systems}

\author{Dongrui Wu and Jerry M. Mendel
\thanks{D.~Wu is with the Key Laboratory of the Ministry of Education for Image Processing and Intelligent Control, School of Artificial Intelligence and Automation, Huazhong University of Science and Technology, Wuhan 430074, China. Email:  drwu@hust.edu.cn.}
\thanks{J. Mendel is with the Ming Hsieh Department of Electrical Engineering, University of Southern California, Los Angeles, CA, USA, and the College of Artificial Intelligence, Tianjin Normal University, Tianjin, China. Email: mendel@sipi.usc.edu.}}

\maketitle

\begin{abstract}
Interval type-2 (IT2) fuzzy systems have become increasingly popular in the last 20 years. They have demonstrated superior performance in many applications. However, the operation of an IT2 fuzzy system is more complex than that of its type-1 counterpart. There are many questions to be answered in designing an IT2 fuzzy system: Should singleton or non-singleton fuzzifier be used? How many membership functions (MFs) should be used for each input? Should Gaussian or piecewise linear MFs be used? Should Mamdani or Takagi-Sugeno-Kang (TSK) inference be used? Should minimum or product $t$-norm be used? Should type-reduction be used or not? How to optimize the IT2 fuzzy system? These questions may look overwhelming and confusing to IT2 beginners. In this paper we recommend some representative starting choices for an IT2 fuzzy system design, which hopefully will make IT2 fuzzy systems more accessible to IT2 fuzzy system designers.
\end{abstract}


\IEEEpeerreviewmaketitle

\section{Introduction} \label{sect:introduction}

\begin{figure}[htpb]\centering
  \subfigure[]{\label{fig:numberTitle} \includegraphics[width=.8\linewidth]{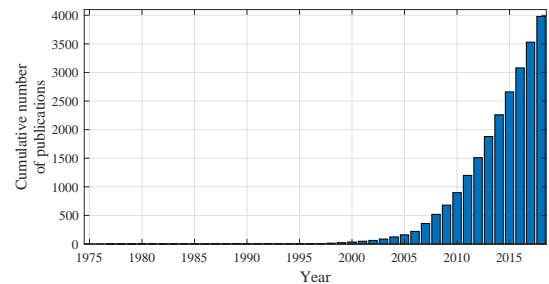}}
  \subfigure[]{\label{fig:numberText} \includegraphics[width=.8\linewidth]{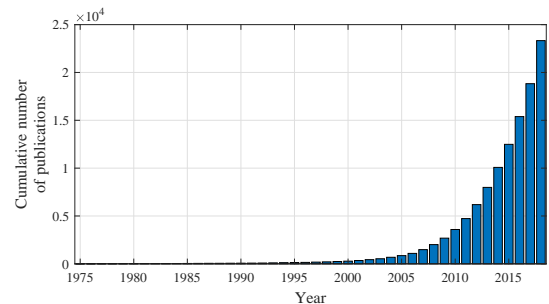}}
\caption{Cumulative number of Google Scholar publications on type-2 fuzzy sets and systems. (a) with ``type-2 fuzzy" in the title. (b) with ``type 2 fuzzy" anywhere in the paper.} \label{fig:pubNumbers}
\end{figure}

Type-2 fuzzy sets were introduced by Zadeh in 1975 \cite{Zadeh1975} but have only become popular during the last 20 years\footnote{More than 1600 articles about type-2 fuzzy sets and systems, including many that are about applications can be found at: https://jmmprof.wixsite.com/jmmprof (click ``Extensive T2 References").}. Fig.~\ref{fig:numberTitle} shows the cumulative number of publications, when searched in Google Scholar using the exact phrase ``type 2 fuzzy" in the title, excluding citations and patents\footnote{We did not count the number of publications about type II and interval-valued fuzzy sets and systems here. The numbers would be larger if we did that.}. Fig.~\ref{fig:numberText} shows the cumulative number of publications, when searched in Google Scholar using the exact phrase ``type 2 fuzzy" anywhere in the paper, excluding citations and patents\footnote{Again, we did not count the number of publications about type II and interval-valued fuzzy sets and systems here.}. Observe that the trends are similar, and both numbers have been increasing quickly since 2000. Another perspective to evaluate the popularity of type-2 fuzzy sets and systems is to look at the awarded Outstanding Papers of the \emph{IEEE Transactions on Fuzzy Systems}\footnote{https://cis.ieee.org/getting-involved/awards/past-recipients\#TFSOutstandingPaperAward}, the flagship journal on fuzzy sets and systems. It has awarded 20 outstanding papers since 2001. Seven of them were on type-2 fuzzy sets and systems, and an eighth one on interval-valued fuzzy sets \cite{Li2010d}, which are closely related to IT2 fuzzy sets \cite{Bustince2016,Mendel2016,Sola2015}. Remarkably, five of them were awarded very recently.

Although type-2 fuzzy sets and systems are gaining popularity, many researchers and practitioners are still using traditional type-1 (T1) fuzzy sets and systems. We argue that type-2 fuzzy sets and systems should be used, for the following reasons \cite{Mendel2019}:
\begin{enumerate}
\item Type-2 fuzzy systems offer some novel partitions of the input domain \cite{Mendel2018}, which cannot be achieved by T1 fuzzy systems.
\item  Type-2 fuzzy sets are the right models for membership function (MF) uncertainties \cite{drwuBook2010}.
\item Using type-2 fuzzy sets to model words is scientifically correct \cite{drwuBook2010,Mendel2007}, whereas using T1 fuzzy sets is not.
\item Type-2 fuzzy systems have demonstrated superior performance in many applications \cite{Castillo2008Book2,Dereli2011,Hagras2007,Hagras2012,drwuIT2FLC2018,Liang20004,drwuEAAI2006}, particularly modeling and control.
\end{enumerate}

Our design philosophy is to first design the best possible T1 fuzzy system. If such a system is unable to provide satisfactory performance, then move up to interval type-2 (IT2) fuzzy systems. This philosophy acknowledges that although IT2 fuzzy systems are simpler than (general) type-2 fuzzy systems, they are still more complex than T1 fuzzy systems. This paper therefore assumes that the designer has decided to use a type-2 fuzzy system instead of a T1 fuzzy system. Particularly, he/she will use an IT2 fuzzy system \cite{Mendel2017}, which has dominated the research and applications of type-2 fuzzy systems so far, due to its simpler structure and reduced computational cost, as compared to a general type-2 fuzzy system.

There are many questions to be answered in designing an IT2 fuzzy system: Should a singleton or non-singleton fuzzifier be used? How many MFs should be used for each input? Should Gaussian or piecewise linear MFs be used? Should Mamdani or Takagi-Sugeno-Kang (TSK) inference be used? Should minimum or product $t$-norm be used? Should output processing include type-reduction\footnote{In an IT2 fuzzy system, type-reduction can be used to reduce IT2 fuzzy sets into T1 fuzzy sets, so that defuzzification can then be performed to generate a crisp output.}? If type-reduction is used, which method should be chosen? How to optimize the IT2 fuzzy system? Of these the most difficult questions are the ones about output processing and type-reduction methods. There are many type-reduction methods. For example, \cite{drwuCCTFS2013} presented six methods to compute the exact type-reduced outputs, as well as 11 alternatives (see, also \cite{Mendel2013}).

While the answers to these questions give an experienced IT2 fuzzy system researcher extensive freedom to design an optimal IT2 fuzzy system, the questions may look overwhelming and confusing to IT2 beginners. Such a beginner may make an inappropriate choice, obtain unexpected results, and lose interest, which will hinder the wider applications of IT2 fuzzy systems. In this paper we try to lower the learning barriers for an IT2 beginner by recommending some arguably representative starting choices for an IT2 fuzzy system design.

The remainder of this paper\footnote{This paper is developed from our conference paper at IEEE WCCI 2014 \cite{drwuPractical2014}. Some statements and conclusions have changed.} is organized as follows: Because the design of IT2 fuzzy system builds upon the experience of designing a T1 fuzzy system, Section~II provides parallel discussions about practical T1 and IT2 fuzzy system designs, and illustrates our recommended choices with an example; Section~III clarifies two myths about IT2 fuzzy systems; and, Section~IV draws conclusions. We assume the readers have some familiarity with both T1 and IT2 fuzzy sets and fuzzy systems, so we will not explain in detail basic concepts like MFs, upper MFs (UMFs), lower MFs (LMFs), and footprint of uncertainty (FOU). These definitions are given in Table~\ref{tab:Notation} and can also be found in \cite{Mendel2017,Mendel2007b}.

\emph{Disclaimer}: The recommendations in this paper are based on the two authors' combined 40+ years of research on IT2 fuzzy systems. We tried our best to keep them up-to-date and unbiased. However, it is still possible that some latest progresses are not reflected, and the recommendations may not always result in the best fuzzy system. They are meant to be good starting points for beginners and practitioners.

\begin{table*}[htpb]\centering
\caption{Notations or definitions for type-2 fuzzy sets.} 
\begin{tabular}{lll} \hline
\multicolumn{1}{c}{Term} & \multicolumn{1}{c}{Literal definition} & \multicolumn{1}{c}{Mathematical definition} \\ \hline
$\tilde{A}$ & General T2 fuzzy set (GT2 fuzzy set) & $\tilde{A}=\left\{\left((x,u),\mu_{\tilde{A}}(x,u)\right)|x\in X, u\in U\equiv [0,1]\right\}$ \\
$\tilde{A}(x)$ & Name of secondary T1 fuzzy set & NA \\
$\mu_{\tilde{A}(x)}(u)$ & Secondary MF [also called  & A restriction of function $\mu_{\tilde{A}}: X \times [0,1]\rightarrow[0,1]$ to $x\in X$, \\
 & a vertical slice of $\mu_{\tilde{A}}(x,u)$] & i.e., $\mu_{\tilde{A}(x)}:[0,1]\rightarrow[0,1]$, or $\mu_{\tilde{A}(x)}(u)=\int_{u\in[0,1]}\mu_{\tilde{A}}(x,u)/u$\\
$J_x$ & Primary membership of $x$ & $J_x=\{(x,u)|u\in[0,1],\mu_{\tilde{A}}(x,u)>0\}$\\
$I_x$ & Support (can be connected or& $I_x=\{u\in[0,1]|\mu_{\tilde{A}}(x,u)>0\}$ so that $J_x=\{x\}\times I_x$\\
& disconnected) of  the secondary MF&\\
$DOU(\tilde{A})$ & Domain of uncertainty & $DOU(\tilde{A})=\{(x,u)\in X\times [0,1]|\mu_{\tilde{A}}(x,u)>0\}=\cup_{x\in X}J_x$ \\ \hline \hline
\multicolumn{3}{c}{Special situation when all secondary grades =1} \\ \hline
\multicolumn{1}{c}{Term} & \multicolumn{1}{c}{Literal definition} & \multicolumn{1}{c}{Mathematical definition} \\ \hline
$\tilde{A}$ & IT2 fuzzy set & $\tilde{A}=\left\{\left((x,u),\mu_{\tilde{A}}(x,u)=1\right)|x\in X, u\in U\equiv [0,1]\right\}$ \\
 \hline \hline
\multicolumn{3}{c}{Special situation when all secondary grades =1 and $I_x$ is closed} \\ \hline
$\tilde{A}$ & Closed IT2 fuzzy set$^a$& $\tilde{A}=\left\{\left((x,u),\mu_{\tilde{A}}(x,u)=1\right)|x\in X, u\in I_x\right\}$ \\
$I_x$ & Closed support of the secondary MF& $I_x=\{u\in[0,1]|\mu_{\tilde{A}}(x,u)>0\}=[\mu_{\underline{A}}(x),\mu_{\overline{A}}(x)]$ \\
$\mu_{\underline{A}}(x)$ & Lower MF of $FOU(\tilde{A})$ & $\mu_{\underline{A}}(x)=\inf \{u|u\in[0,1],\mu_{\tilde{A}}(x,u)>0\}$\\
$\mu_{\overline{A}}(x)$ & Upper MF of $FOU(\tilde{A})$ & $\mu_{\overline{A}}(x)=\sup\{u|u\in[0,1], \mu_{\tilde{A}}(x,u)>0\}$\\
$FOU(\tilde{A})$ & Footprint of uncertainty & $DOU(\tilde{A})=FOU(\tilde{A})=\{(x,u)|x\in X, u\in[\mu_{\underline{A}}(x),\mu_{\overline{A}}(x)]\}$\\ \hline \hline
\multicolumn{3}{l}{\small $^a$ Most articles about IT2 fuzzy sets and systems use closed IT2 fuzzy sets. When it is clear that this is the situation,} \\
\multicolumn{3}{l}{\small then ``closed IT2 fuzzy set"  can be replaced by ``IT2 fuzzy set".}
  \end{tabular}   \label{tab:Notation}
\end{table*}

\section{Considerations for Practical T1 and IT2 Fuzzy System Designs}

\begin{figure}[htpb]\centering
  \subfigure[]{\label{fig:2a} \includegraphics[width=.8\linewidth]{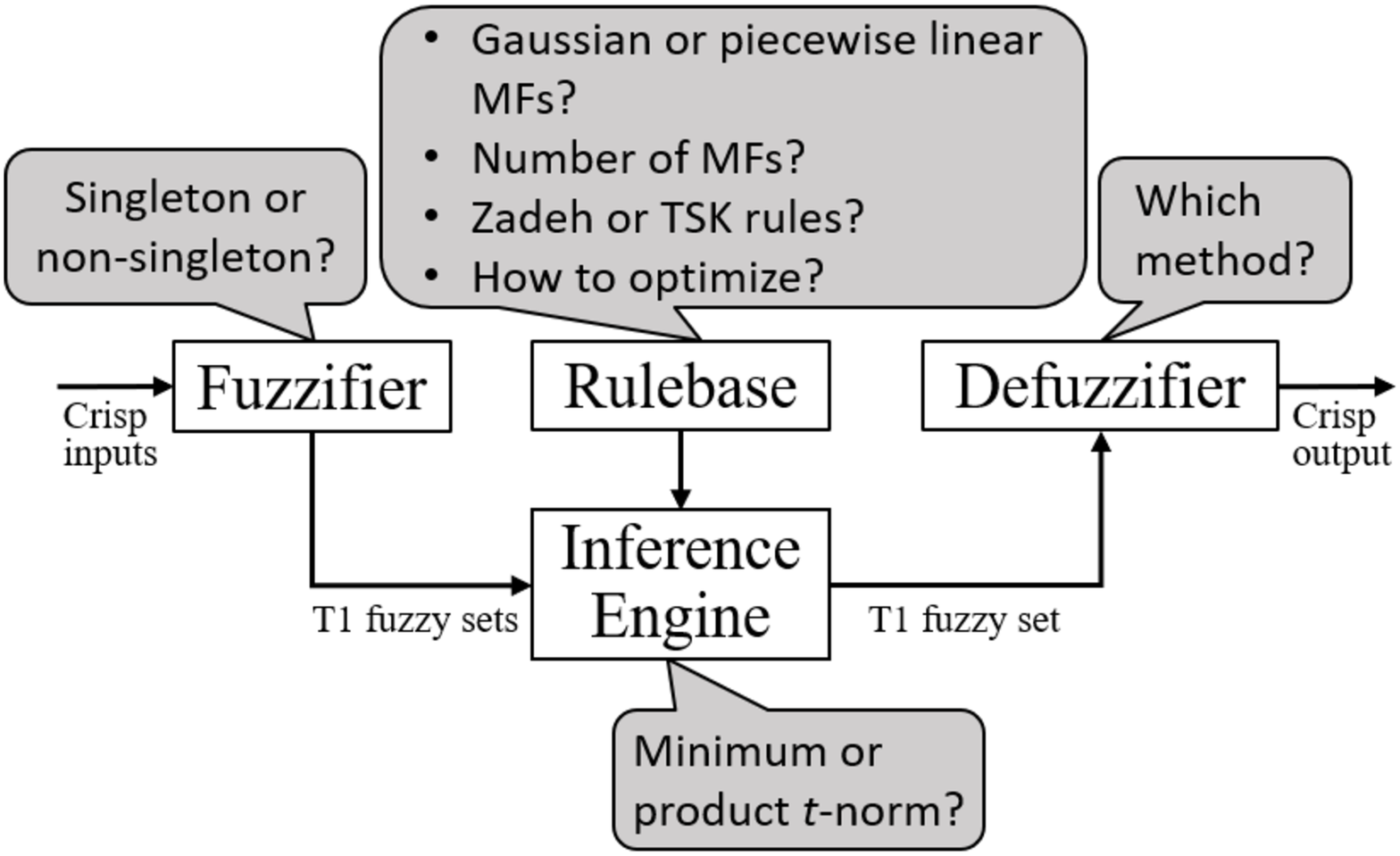}}\hfill
  \subfigure[]{\label{fig:2b} \includegraphics[width=.9\linewidth]{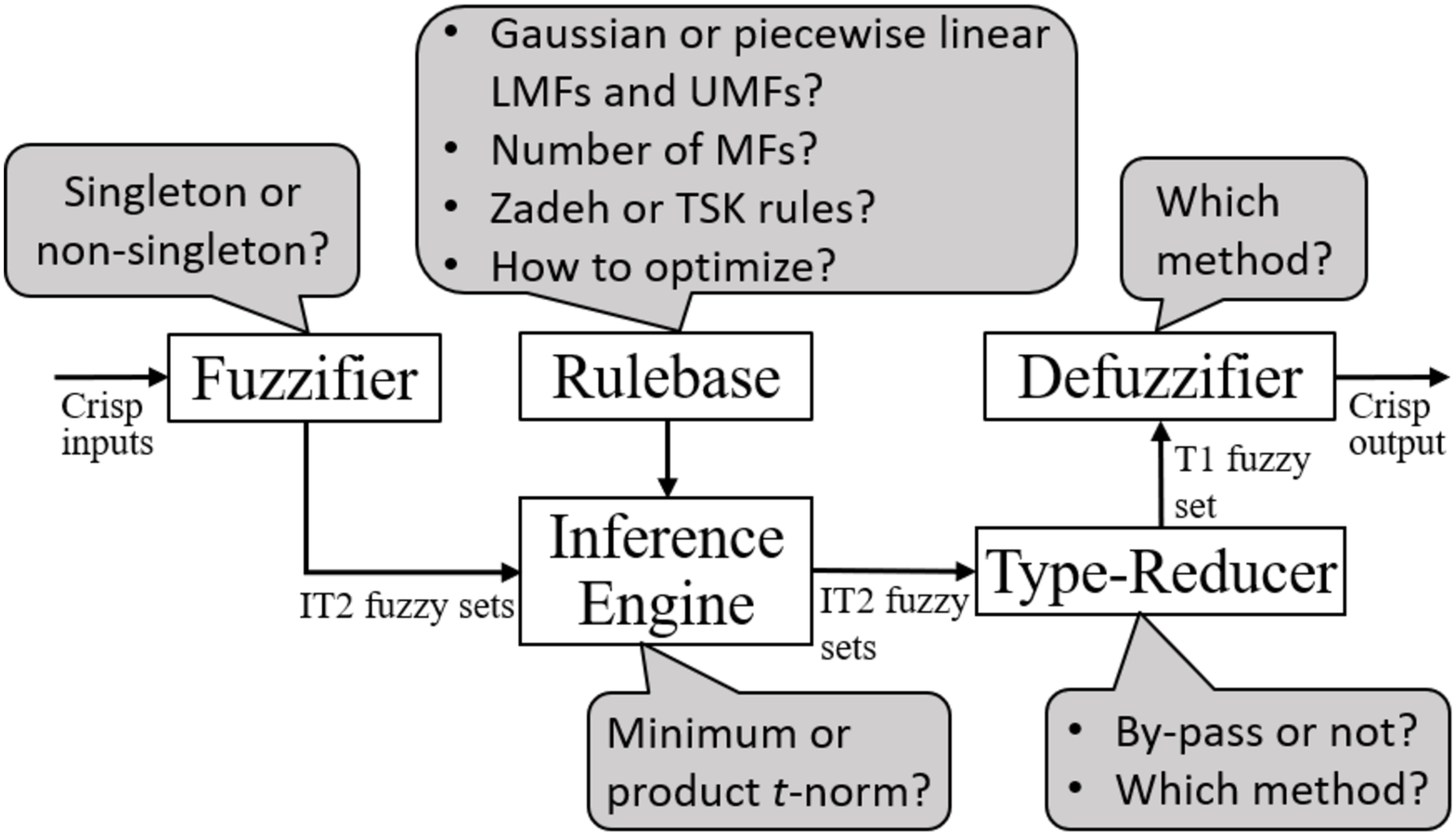}}
\caption{Structures of and design choices for (a) T1 fuzzy system and (b) IT2 fuzzy system. } \label{fig:2}
\end{figure}

A diagram of a T1 fuzzy system is shown in Fig.~\ref{fig:2a}. It consists of four components: \emph{fuzzifier}, \emph{rulebase}, \emph{inference engine}, and \emph{defuzzifier}. As shown in the annotation boxes, there are many choices to be made in a practical T1 fuzzy system design.

The diagram of an IT2 fuzzy system is shown in Fig.~\ref{fig:2b}. Its rules use IT2 fuzzy sets instead of T1 fuzzy sets; as a result, it may need an extra step called \emph{type-reduction} before the defuzzifier to reduce IT2 fuzzy sets into T1 fuzzy sets. Observe that there are even more choices to be made in practical IT2 fuzzy system designs, as shown in the annotation boxes in Fig.~\ref{fig:2b}.

Type-1 and IT2 fuzzy systems (with input $\mathbf{x}\equiv [x_1;\ldots;x_p]$) are nonlinear variable-structure systems with output $y=f(\mathbf{x})$, meaning that they automatically partition the state space $X_1\times\cdots \times X_p$ into a multitude of nonlinear subsystems (i.e., they sculpt the state space). This occurs by virtue of their overlapping MFs of the linguistic terms of the $p$ inputs.

\cite{Mendel2018} explains and demonstrates how a T1 fuzzy system can sculpt its state space with greater variability than a crisp rule-based system can, and in ways that cannot be accomplished by the crisp system, and that an IT2 fuzzy system (that has the same number of rules as the T1 fuzzy system) can sculpt the state space with even greater variability, and in ways that cannot be accomplished by a T1 fuzzy system. It is then conjectured that it is the greater sculpting of the state space by a T1 fuzzy system that lets it outperform a crisp system, and it is the even greater sculpting of the state space by an IT2 fuzzy system that lets it outperform a T1 fuzzy system (the latter can occur even when the T1 and IT2 fuzzy systems are described by the same number of parameters).

In the rest of this section we describe the most important practical design considerations for both T1 and IT2 fuzzy systems, which should be helpful especially to IT2 beginners.

\subsection{Fuzzifier: Singleton or Non-Singleton}

\begin{figure*}[htpb]\centering
\includegraphics[width=.8\linewidth]{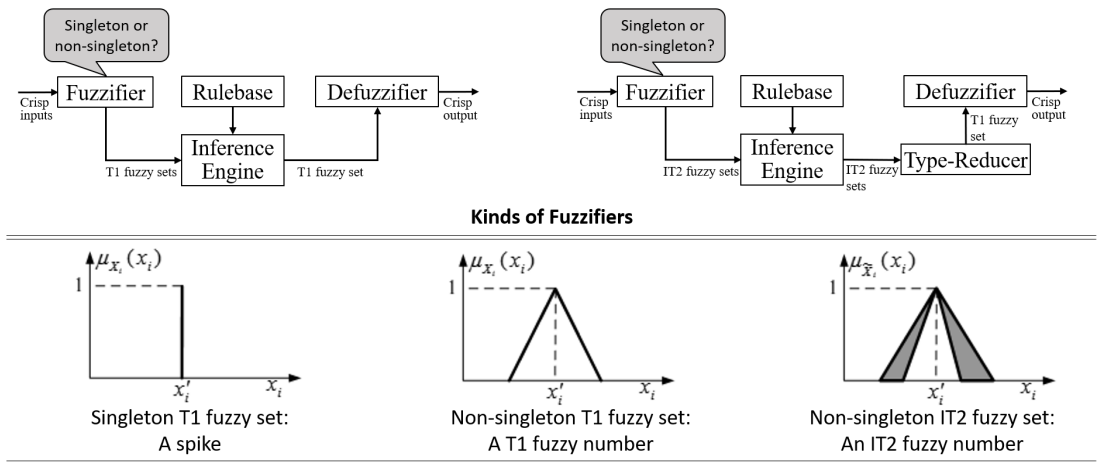} \caption{Fuzzifiers and their choices: two choices for a T1 fuzzy system (a spike or a T1 fuzzy number) and three choices for an IT2 fuzzy system (a spike or a T1 fuzzy number or an IT2 fuzzy number).} \label{fig:3}
\end{figure*}

The fuzzifier of a T1 fuzzy system maps an input vector $\mathbf{x}=(x_1',...,x_p')^T$ into $p$ T1 fuzzy sets $X_i$, $i=1,2,...,p$. There are two categories of fuzzifiers \cite{Mendel2017} for a T1 fuzzy system: \emph{singleton} and \emph{non-singleton}. For a singleton fuzzifier, $\mu_{X_i}(x_i)=1$ at $x_i=x_i'$ and $\mu_{X_i}(x_i)=0$ everywhere else, as shown in the first subfigure in the bottom row of Fig.~\ref{fig:3}. For a non-singleton fuzzifier, $\mu_{X_i}(x_i)=1$ at $x_i=x_i'$ and $\mu_{X_i}(x_i)$ decreases from unity as $x_i$ moves away from $x_i'$, as shown in the middle subfigure in the bottom row of Fig.~\ref{fig:3}. Conceptually, the non-singleton fuzzifier implies that the given input value $x_i'$ is the most likely value to be the correct one from all the values in its immediate neighborhood; however, because the input is corrupted by noise, neighboring points are also likely to be the correct value, but to a lesser degree \cite{Mendel2017}. Usually non-singleton $\mu_{X_i}(x_i)$ is symmetric about $x_i'$ because the effect of noise is most likely to be equivalent on all points. A non-singleton fuzzifier can be thought of as a pre-filter of $x_i'$.

Similarly, the fuzzifier of an IT2 fuzzy system maps an input vector $\mathbf{x}=(x_1',...,x_p')^T$ into $p$ IT2 fuzzy sets $\tilde{X}_i$, $i=1,2,...,p$. Like its T1 counterpart, the fuzzifier of an IT2 fuzzy system can also be singleton or non-singleton \cite{Mendel2017}. For a singleton fuzzifier, $\mu_{\tilde{X}_i}(x_i)=1/1$ (both the primary and secondary memberships are 1) at $x_i=x_i'$ and $\mu_{\tilde{X}_i}(x_i)=1/0$ (the primary membership is 0, and secondary membership is 1) everywhere else, as shown in the first subfigure in the bottom row of Fig.~\ref{fig:3}. For a non-singleton fuzzifier, the output can be a T1 fuzzy set, as shown in the middle subfigure in the bottom row of Fig.~\ref{fig:3}, or even an IT2 fuzzy set, as shown in the last subfigure in the bottom row of Fig.~\ref{fig:3}. The latter is very useful when the measurements are corrupted by non-stationary noise. Indeed, non-singleton fuzzifiers have demonstrated better performance than singleton fuzzifiers in some such applications \cite{Chua2007,Mendel2017}.

Our recommendation is that, when the measurements are very noisy, non-singleton fuzzification should be considered to accommodate them; otherwise, one should begin with singleton fuzzification for both T1 and IT2 fuzzy systems, which is much more popular in practice due to its simplicity. We will only consider singleton fuzzy systems in the sequel.

\subsection{Rulebase: Gaussian or Piecewise Linear MFs?}

\begin{figure*}[htpb]\centering
\includegraphics[width=.9\linewidth]{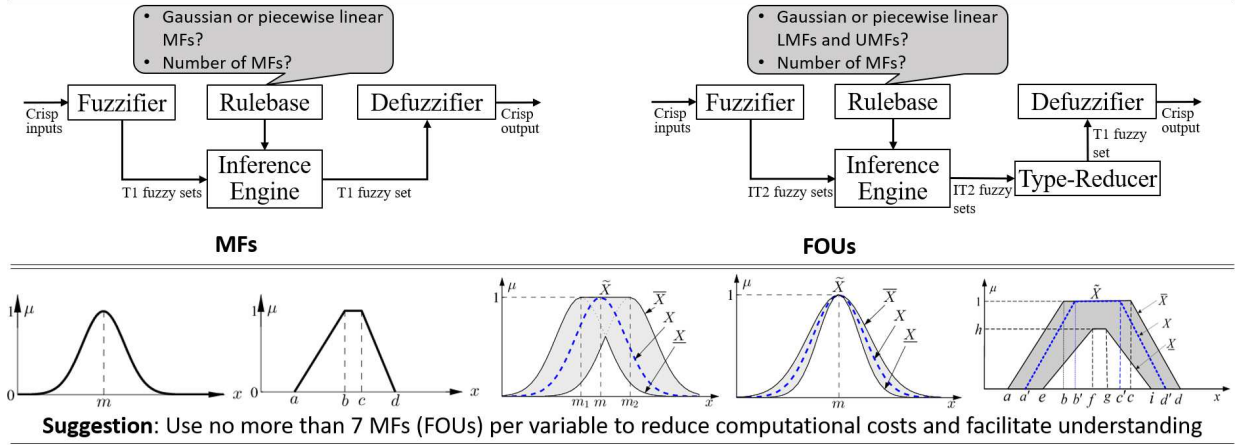} \caption{Rulebase: Gaussian or piecewise linear MFs or FOUs? How many?} \label{fig:4}
\end{figure*}

The two most commonly used MF shapes for T1 fuzzy systems are Gaussian and piecewise linear. A Gaussian T1 fuzzy set is shown in the first subfigure in the bottom row of Fig.~\ref{fig:4}, and is described by
\begin{align}
\mu(x)=e^{-\frac{(x-m)^2}{2\sigma^2}}, \label{eq:T1G}
\end{align}
where $m$ determines the center and $\sigma$ determines the spread.

A piecewise linear T1 fuzzy set is shown in the second subfigure in the bottom row of Fig.~\ref{fig:4}. The most popular piecewise linear MF shape is trapezoidal, determined by four parameters $(a,\,b,\,c,\,d)$:
\begin{align}
\mu(x)=\left\{\begin{array}{ll}
                  \frac{x-a}{b-a}, & a< x < b\\
                  1, & b\le x \le c \\
                  \frac{d-x}{d-c}, & c<x<d\\
                  0, & \mbox{otherwise}
                \end{array}\right.. \label{eq:T}
\end{align}
Note that a triangular T1 fuzzy set is a special case of a trapezoidal T1 fuzzy set when $b=c$. Comparing (\ref{eq:T1G}) and (\ref{eq:T}), observe that (\ref{eq:T}) requires tests about the location of its independent variable, but (\ref{eq:T1G}) does not. This complicates derivative computations (performed during training) when (\ref{eq:T}) is used, and may make (\ref{eq:T}) more costly to use in real-time applications, although the numerical calculations of (\ref{eq:T1G}) may also be as costly.

Performance is the most important consideration in choosing between Gaussian and piecewise linear MFs in a T1 fuzzy system, and different applications of T1 fuzzy systems have different definitions of performance. The most popular application is fuzzy control. There are many studies on comparing the control performance of Gaussian and piecewise linear MFs in T1 fuzzy controllers \cite{Gupta2010,Monicka2011}; however, it seems that the conclusion is highly problem dependent, and it is difficult to conclude which MF shape is always better for control performance. We expect the conclusion will be the same for other applications of T1 fuzzy systems, including classification, regression, etc. So, we do not have a preference on the shape of MFs in a T1 fuzzy system in terms of performance. However, we need to point out that the input-output mapping of a T1 fuzzy system may have discontinuities if the input MFs do not cover each input domain completely \cite{drwuCont2011}.  This is an advantage of Gaussian MFs over piecewise linear MFs because the former always spread out over the entire input domains.

The FOU in an IT2 fuzzy set also has two main categories of shapes: Gaussian and piecewise linear. A Gaussian IT2 fuzzy set is usually obtained by blurring the mean or standard deviation of a baseline Gaussian T1 fuzzy set \cite{drwuEAAI2006}, as shown in the third and fourth subfigures in the bottom row of Fig.~\ref{fig:4}. In either case, only three parameters [$(m_1,m_2,\sigma)$ or $(m,\sigma_1,\sigma_2)$] are needed to define a Gaussian IT2 fuzzy set.

When the mean of the Gaussian T1 fuzzy set is blurred to be an interval $[m_1,\, m_2]$, as shown in the third subfigure in the bottom row of Fig.~\ref{fig:4}, the UMF is
\begin{align}
\mu_{\overline{X}}(x)=\left\{\begin{array}{ll}
                                 e^{-\frac{(x-m_1)^2}{2\sigma^2}}, & x<m_1 \\
                                 1, & m_1\le x\le m_2 \\
                                 e^{-\frac{(x-m_2)^2}{2\sigma^2}}, & x>m_2 \\
                               \end{array}\right.
\end{align}
and the LMF is
\begin{align}
\mu_{\underline{X}}(x)=\min\left(e^{-\frac{(x-m_1)^2}{2\sigma^2}},e^{-\frac{(x-m_2)^2}{2\sigma^2}}\right).
\end{align}
When the standard deviation of the Gaussian T1 fuzzy set is blurred to be an interval $[\sigma_1,\, \sigma_2]$, as shown in the fourth subfigure in the bottom row of Fig.~\ref{fig:4}, the UMF is
\begin{align}
\mu_{\overline{X}}(x)=e^{-\frac{(x-m)^2}{2\sigma_2^2}}
\end{align}
and the LMF is
\begin{align}
\mu_{\underline{X}}(x)=e^{-\frac{(x-m)^2}{2\sigma_1^2}}.
\end{align}
From the above formulas it seems that the memberships of a Gaussian FOU with uncertainty standard deviations are easier to compute than a Gaussian FOU with uncertain means, which may offer the former a slight advantage in implementation. However, note that this does not mean the former also has better performance than the latter. To our knowledge, there has not been a comprehensive study and definite conclusion on this.

Of course, one can also blur both the mean and the standard deviation to obtain a more general Gaussian FOU, but this approach is rarely used in practice.

A piecewise linear IT2 fuzzy set can also be obtained by blurring a baseline piecewise linear T1 fuzzy set, as shown in the last subfigure in the bottom row of Fig.~\ref{fig:4}. Generally, nine parameters are needed to represent a piecewise linear IT2 fuzzy set, $(a,\,b,\,c,\,d,\,e,\,f,\,g,\,i,\,h)$ shown in that subfigure, where $(a,\,b,\,c,\,d)$ determines the UMF and $(e,\,f,\,g,\,i,\,h)$ determines the sub-normal LMF. The UMF is still computed by (\ref{eq:T}), and the LMF is computed by
\begin{align}
\mu_{\underline{X}}(x)=\left\{\begin{array}{ll}
           h\cdot\frac{x-e}{f-e}, & e< x < f\\
                  h, & f\le x \le g \\
                  h\cdot\frac{i-x}{i-g}, & g<x<i\\
                  0, & \mbox{otherwise}
                \end{array}\right..
\end{align}

From the above description we can conclude that generally it is simpler to represent a Gaussian IT2 fuzzy set because it only needs three or four parameters, whereas a piecewise linear IT2 fuzzy set needs nine parameters.

\begin{figure*}[htpb]\centering
\subfigure[]{\label{fig:T2MFs}\includegraphics[width=.8\linewidth]{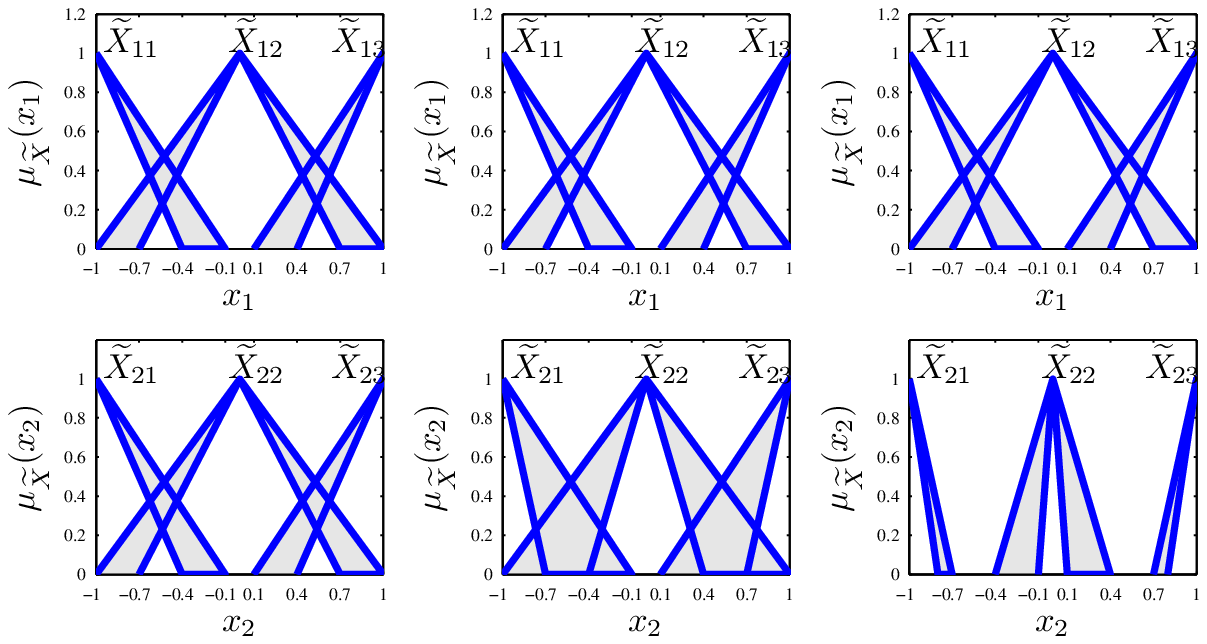}}
\subfigure[]{\label{fig:T2fuzzy system2KM}\includegraphics[width=\linewidth]{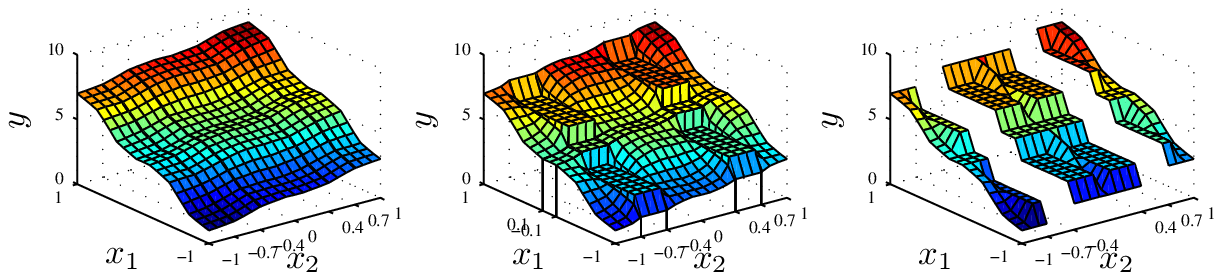}}
\caption{Example input-output mappings of 2-input IT2 fuzzy systems. (a) The input MFs; (b) The input-output mappings computed by the center-of-sets type-reducer.} \label{fig:T2fuzzy system2}
\end{figure*}

In \cite{drwuMFs2012} we presented 12 considerations about choosing between Gaussian and piecewise linear FOUs for an IT2 fuzzy system, including representation, construction, optimization, adaptiveness, novelty, analytical structure, continuity, monotonicity, stability, robustness, computational cost, and control performance. Here we focus only on continuity because it was widely ignored before \cite{drwuCont2011}. The following example illustrates the input-output mappings of three 2-input IT2 fuzzy systems using the popular center-of-sets type-reducer \cite{Karnik2001,Mendel2013}, computed by the EIASC algorithms introduced in Section~\ref{sect:TR}. Fig.~\ref{fig:T2MFs} shows the three FOUs in each input domain, and Fig.~\ref{fig:T2fuzzy system2KM} shows the corresponding input-output mappings. The FOUs for $x_1$ are the same in all the cases, whereas the FOUs for $x_2$ are not. Observe that:
\begin{enumerate}
\item When the input UMFs and LMFs for both $x_1$ and $x_2$ fully cover their domains\footnote{The LMFs for $x$ fully cover its domain means the sum of membership degrees on the LMFs for any $x$ is larger than 0. Because an UMF is always above or equal to the corresponding LMF, when LMFs cover an input domain, it is automatically guaranteed that the UMFs also cover the input domain.}, as shown in the first column of Fig.~\ref{fig:T2MFs}, the corresponding input-output mapping is continuous.
\item When the two input domains are fully covered by the UMFs but at least one point in the domain of $x_2$ is not covered by the LMFs, as shown in the middle column of Fig.~\ref{fig:T2MFs}, the corresponding input-output mapping has jump discontinuities (at $x_2=\{\pm0.4,\pm0.7\}$).
\item When the input UMFs and LMFs for $x_2$ do not fully cover its domain, as shown in the last column of Fig.~\ref{fig:T2MFs}, the corresponding input-output mapping has both gap discontinuities and jump discontinuities.
\end{enumerate}

Gaussian IT2 fuzzy sets are frequently used in IT2 fuzzy systems. Mendel \cite{Mendel2018} summarized three reasons: 1) they guarantee the continuity of the resulting IT2 fuzzy system, as described above; 2) their derivatives with respect to their parameters are easier to compute in gradient-based optimization algorithms; and, 3) they are emphasized in \cite{Mendel2001}, the highest referenced book on type-2 fuzzy systems. Mendel \cite{Mendel2018} also explained, from the viewpoint of sculpting the state space, why one may not see as much performance improvement of IT2 fuzzy systems over T1 fuzzy systems when Gaussian MFs are used in both (however, this does not mean the performance of a Gaussian IT2 fuzzy system is always worse than that of an IT2 fuzzy system with piecewise linear FOUs).

Piecewise linear IT2 fuzzy sets are also very popular in practice. Comparing with their Gaussian counterparts, analytical structures of IT2 fuzzy systems  with piecewise linear FOUs are much easier to derive \cite{Du2010,Nie2012,drwuFundamental2012,Zhou2013,Zhou2017}, though still very complex. And, Mendel's \cite{Mendel2001} uncertainty partition, rule-partition, and novelty partition results, which explain the performance potential of fuzzy systems as a greater sculpting of the state space, mainly apply to IT2 fuzzy systems with piecewise linear FOUs.

In summary, each kind of IT2 fuzzy systems has its own advantages: Gaussian IT2 fuzzy systems are simpler in design because they are easier to represent and to optimize, always continuous, and faster to compute for small rulebases, whereas IT2 fuzzy systems with piecewise linear FOUs are easier to analyze. We recommend Gaussian MFs for T1 fuzzy systems and Gaussian FOUs for IT2 fuzzy systems, for their simplicity and automatic guarantee of continuity.

\subsection{Rulebase: How Many MFs?}

The next natural question on rulebase design is how many MFs should be used in each input domain. Theoretically, there is no constraint on the number of MFs one could use; however, in practice some considerations may prevent one from using too many MFs. First, because the number of rules is an exponential function of the number of MFs in each input domain (e.g., for a 2-input fuzzy system, if each input domain consists of 3 MFs, then the total number of rules is $3^2=9$; however, if each input domain consists of 9 MFs, then the total number of rules becomes $9^2=81$), the computational cost increases rapidly with the number of MFs. Second, some people may prefer fuzzy systems to other black-box models, e.g., neural networks, because fuzzy systems can be interpreted by looking at the rules; however, this advantage diminishes as the number of rules increases.

It is well-known in psychology that the number of objects an average human can hold in working memory is $7\pm 2$ \cite{Miller1956}. We also suggest $\le7$ MFs in each input domain of a T1 or IT2 fuzzy system to reduce computational cost and to facilitate interpretation.

\subsection{Rulebase: Zadeh or TSK Rules}

\begin{figure*}[htpb]\centering
\includegraphics[width=.8\linewidth]{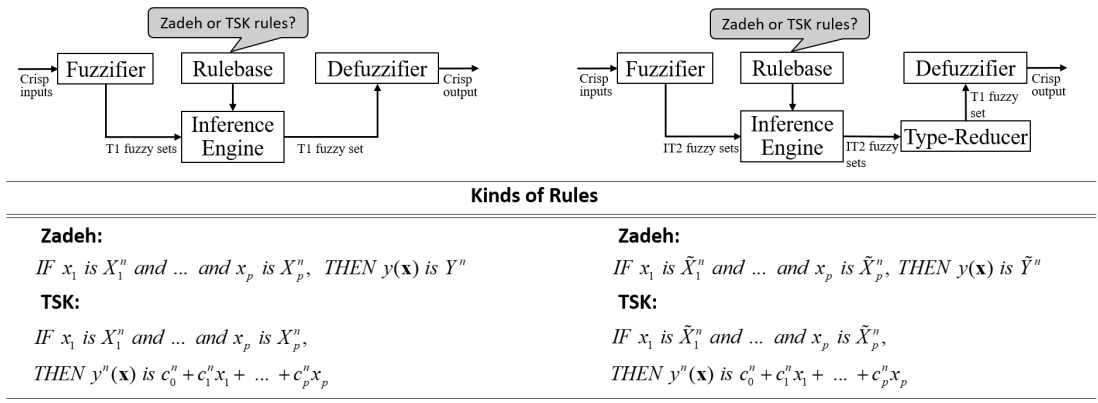} \caption{Rulebase: Zadeh or TSK rules?} \label{fig:5}
\end{figure*}

There are two kinds of rules for a T1 fuzzy system: \emph{Zadeh} \cite{Chang1972}, where the rule consequents are fuzzy sets, and \emph{TSK} \cite{Takagi1985}, where the rule consequents are crisp functions of the inputs. As shown in Fig.~\ref{fig:5}, for a $p$-input T1 fuzzy system, a Zadeh rule is of the form:
\begin{align*}
R^n:\quad &\mbox{IF } x_1 \mbox{ is } X_1^n \mbox{ and }\ldots \mbox{ and } x_p  \mbox{ is } X_p^n,\\
 &\mbox{THEN } y(\mathbf{x})\mbox{ is } Y^n, \quad n=1,\ldots,N
\end{align*}
where $\mathbf{x}=(x_1,\ldots,x_p)$, and $Y^n$ is a T1 fuzzy set. A TSK rule is of the form:
\begin{align*}
R^n:\quad &\mbox{IF } x_1 \mbox{ is } X_1^n \mbox{ and }\ldots \mbox{ and } x_p  \mbox{ is } X_p^n,\\
 &\mbox{THEN } y(\mathbf{x})=c_0^n+c_1^n\cdot x_1 +\cdots+c_p^n\cdot x_p, 
\end{align*}
where $\{c_k^n\}_{k=0,\ldots,p}$ are crisp coefficients. Note that more complicated nonlinear functions can also be used in the consequent of a TSK rule.

Zadeh rules were the earliest rules proposed. A popular approach for constructing such rules from data is the Wang-Mendel method \cite{Wang1992}. However, TSK rules are much more popular in practice due to their simplicity and flexibility. A popular approach for constructing such rules from data is the adaptive-network-based fuzzy inference system (ANFIS) \cite{Jang1993}, which has been implemented in the Matlab Fuzzy Logic Toolbox. In many applications people set $c_k^n=0$ $\forall k\in[1,p]$ and each TSK rule consequent is simply represented by a number $c_0^n$. As will be shown later in this section, under certain popular defuzzification methods Zadeh rules are equivalent to the simplest TSK rules; so, we suggest starting from the simplest TSK rules directly for a T1 fuzzy system.

There are also two kinds of rules for an IT2 fuzzy system: \emph{Zadeh}, where the rule consequents are IT2 fuzzy sets, and \emph{TSK}, where the rule consequents are crisp functions of the inputs. For example, for a $p$-input IT2 fuzzy system, a Zadeh rule is of the form
\begin{align}
\tilde{R}^n:\quad &\mbox{IF } x_1 \mbox{ is } \tilde{X}_1^n \mbox{ and }\ldots \mbox{ and } x_p  \mbox{ is } \tilde{X}_p^n,\nonumber\\
 &\mbox{THEN } y(\mathbf{x}) \mbox{ is } \tilde{Y}^n, \quad n=1,\ldots,N \label{eq:Rule}
\end{align}
where $\tilde{Y}^n$ is an IT2 fuzzy set. A TSK rule is of the form
\begin{align*}
\tilde{R}^n:\quad &\mbox{IF } x_1 \mbox{ is } \tilde{X}_1^n \mbox{ and }\ldots \mbox{ and } x_p  \mbox{ is } \tilde{X}_p^n, \\
&\mbox{THEN } y^n(\mathbf{x})=[\underline{y}^n,\overline{y}^n]\\
&\hspace*{19mm}=[\underline{c}_0^n+\underline{c}_1^n x_1 +\cdots+\underline{c}_p^n x_p, \\
&\hspace*{25mm} \overline{c}_0^n+\overline{c}_1^n x_1 +\cdots+\overline{c}_p^n x_p]
\end{align*}
where $\underline{y}^n$, $\overline{y}^n$, $\{\underline{c}_k^n\}_{k=0,\ldots,p}$ and $\{\overline{c}_k^n\}_{k=0,\ldots,p}$ are crisp numbers. More complicated nonlinear functions can also be used in the consequent of the above TSK rule. For simplicity, one can set $\underline{c}_k^n=\overline{c}_k^n=c_k^n$ $\forall k\in[1,\ p]$ and $\forall n\in[1,\ N]$, in which case each rule consequent becomes a single function of the inputs instead of an interval of functions (this is the situation shown in the bottom panel (rhs) of Fig.~\ref{fig:5}). One can also set $\underline{c}_k^n=\overline{c}_k^n=0$ $\forall k\in[1,\ p]$ and $\forall n\in[1,\ N]$, in which case each rule consequent becomes a constant interval $[\underline{c}_0^n,\ \overline{c}_0^n]$. In the simplest case, one sets $\underline{c}_0^n=\overline{c}_0^n=c_0^n$, and $\underline{c}_k^n=\overline{c}_k^n=0$ $\forall k\in[1,\ p]$ and $\forall n\in[1,\ N]$, i.e., each rule consequent becomes a single number $c_0^n$. The latter two approaches are much more popular in practice due to their simplicity, and are our recommended forms to start with.

There have also been efforts to extend ANFIS from T1 fuzzy systems to IT2 fuzzy systems \cite{Chen2016,Chen2017a}. The authors used the center-of-sets type-reducer \cite{Karnik2001,Mendel2013}; however, due to a subtle problem involving the mismatch of the switch points, the solution was not optimal. Maybe a direct defuzzification approach that does not involve the switch points, e.g., those introduced in Section~\ref{sect:def}, could be used to remedy this problem. This is a direction that we are currently working on.


\subsection{Inference: Minimum or Product $t$-Norm}

$t$-norms are used by the inference engine of a T1 fuzzy system to combine the firing levels from multiple antecedents. The two most popular $t$-norms are the \emph{minimum} and the \emph{product}. Assume singleton fuzzification is used, a rule has two antecedents, and their firing levels are $\mu(x_1')$ and $\mu(x_2')$, respectively. Then, the firing level of the rule is $\min[\mu(x_1'),\mu(x_2')]$ for the minimum $t$-norm, and $\mu(x_1')\cdot \mu(x_2')$ for the product $t$-norm.

\begin{figure*}[htpb]\centering
\includegraphics[width=.8\linewidth]{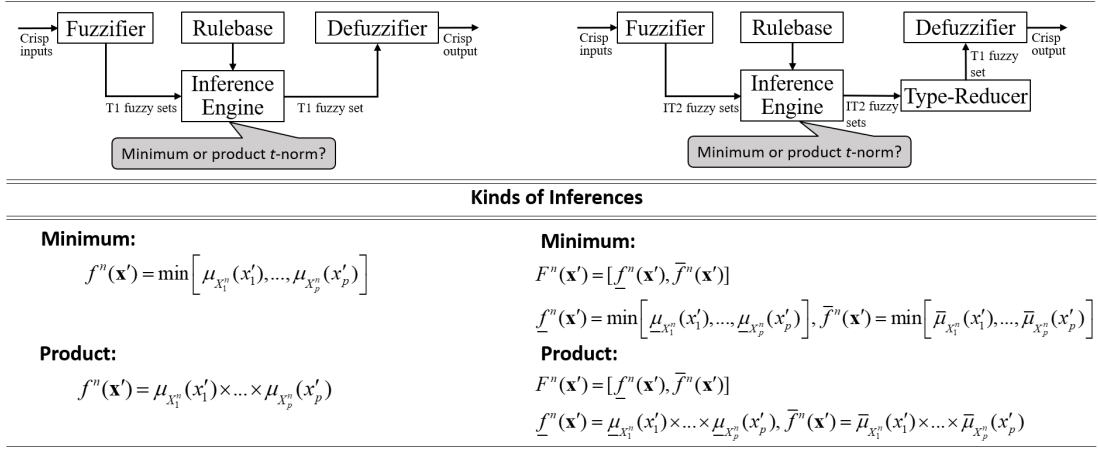} \caption{Inference: Minimum or product used to compute firing level or interval? Formulas are stated for singleton fuzzification.} \label{fig:6}
\end{figure*}

Similarly, minimum and product $t$-norms are also the most popular inference methods to compute firing intervals of the rules for IT2 fuzzy systems. The left (right) endpoint of the interval is computed by applying the corresponding $t$-norm on the LMFs (UMFs). The detailed formulas are giving in Fig.~\ref{fig:6}.

Both $t$-norms have been used extensively for T1 and IT2 fuzzy systems, and there is no evidence that one $t$-norm is better than the other. However, when the product $t$-norm is used, the derivative computation in gradient-based optimization algorithms is generally easier (because taking the minimum of two quantities includes a test, e.g., $\min(a,b)=a$ when $a\le b$, and $\min(a,b)=b$ when $a>b$), and the derivation of the analytical structure of the IT2 fuzzy system is also significantly simpler \cite{Zhou2013}. So, the product $t$-norm is recommended for both T1 and IT2 fuzzy systems.

\subsection{Output Processing}

\begin{figure*}[htpb]\centering
\includegraphics[width=.8\linewidth]{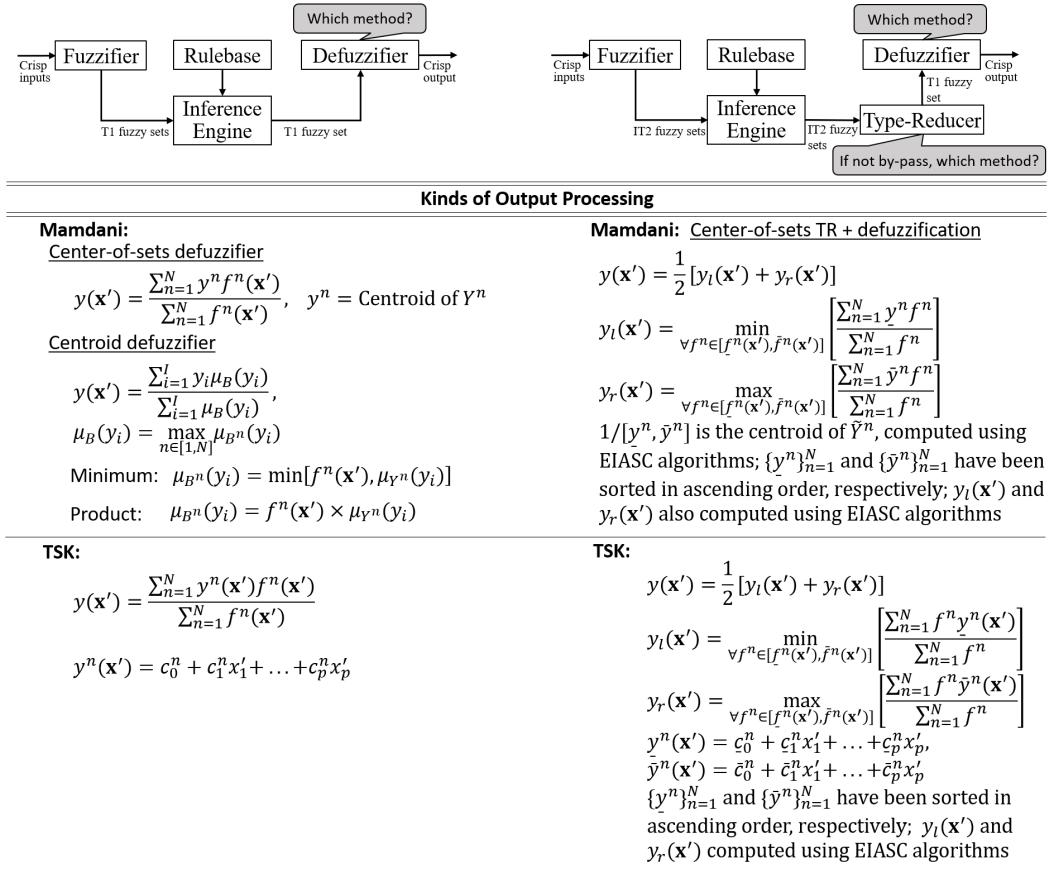} \caption{Output Processing for Mamdani and TSK architectures.} \label{fig:7}
\end{figure*}

How to compute the output of a T1 fuzzy system depends heavily on the kind of rules used. For TSK rules, the computation is straightforward: the output is a weighted average of the crisp rule consequents, where the weights are the firing levels of the rules. The formulas are given in Fig.~\ref{fig:7}.

There are several different methods for computing the output of a Mamdani T1 fuzzy system, which uses Zadeh rules. The formulas are also given in Fig.~\ref{fig:7}. In the center-of-sets defuzzifier, each rule consequent is first replaced by a crisp number, and then a weighted average is used to combine these numbers. In these cases a Mamdani T1 fuzzy system can be viewed as a TSK T1 fuzzy system in which the rule consequents are constants, i.e., $c_k^n=0$ $\forall k\in[1,\ p]$ and $\forall n\in[1,\ N]$. Another defuzzifier used in the early days of Mamdani T1 fuzzy systems is the centroid defuzzifier, which first combines the output T1 fuzzy sets using union and then finds its centroid. Its complexity, due to computing the union, has significantly limited its adoption.

For simplicity and flexibility, we suggest using TSK rules and weighted average to compute the output of a T1 fuzzy system. This is also the choice in the popular ANFIS approach \cite{Jang1993}.

How to compute the output of an IT2 fuzzy system also depends heavily on the kinds of rules used. There are several different methods for computing the output of Mamdani IT2 fuzzy systems \cite{Mendel2017}, which use Zadeh rules. The most popular method uses center-of-sets type-reduction, in which the centroid of each rule consequent IT2 fuzzy set is computed to replace the actual FOU. This is equivalent to the simplified TSK IT2 model, where each rule consequent is an interval $[\underline{c}_0,\ \overline{c}_0]$. This is also our recommended approach for IT2 fuzzy systems.

Next we describe two ways to compute the output in this case for IT2 fuzzy systems.

\subsubsection{Type-Reduction and Defuzzification} \label{sect:TR}

The classical IT2 fuzzy system, as shown in the first row of ``Kinds of Output Processing" in Fig.~\ref{fig:7}, has separate type-reduction and defuzzification steps.

Type-reduction combines $F^n(\mathbf{x}')$, the firing interval of the rules, and $Y^n$, the corresponding rule consequents, to form a T1 fuzzy set $1/[y_l(\mathbf{x'}),\ y_r(\mathbf{x'})]$. There are many type-reduction methods \cite{Mendel2017}, but the most commonly used one is the center-of-sets type-reducer \cite{Karnik2001,Mendel2017}, whose formula is given in Fig.~\ref{fig:7}.

Several efficient methods have been proposed for computing $y_l(\mathbf{x'})$ and $y_r(\mathbf{x'})$ \cite{Duran2008,Hu2012,Hu2010,Karnik2001,Melgarejo2007,drwuEKM2009,Yeh2011,Chen2018a}, including the well-known Karnik-Mendel (KM) algorithms \cite{Karnik2001,Mendel2017}. Comprehensive descriptions and comparisons of the methods are given in  \cite{Mendel2013,drwuCCTFS2013,drwuDA2018}. The speeds of the algorithms are programming language dependent \cite{drwuDA2018}. The Enhanced Iterative Algorithm with Stop Condition (EIASC) \cite{drwuCCTFS2013}, presented in Table~\ref{tab:EIASC}, is the fastest in Matlab, C and Java, whereas the Enhanced KM algorithms \cite{drwuEKM2009} and the optimized direct approach (DA$^*$) \cite{drwuDA2018} are the fastest in R and Python. We recommend the EIASC for its speed and simplicity.

\begin{table}[htbp] \centering  \setlength{\tabcolsep}{0.2cm} \caption{The EIASC \cite{drwuCCTFS2013}. Note that $\{\underline{y}^n\}_{n=1,...,N}$ and $\{\overline{y}^n\}_{n=1,...,N}$ must be sorted in ascending order, respectively.} \label{tab:EIASC}
\begin{tabular}{c|ll}\hline
Step & \multicolumn{1}{|c}{For computing $y_l$} & \multicolumn{1}{c}{For computing $y_r$} \\ \hline
1 & Initialize & Initialize \\
& \quad $a=\sum_{n=1}^N\underline{y}^n\underline{f}^n$ & \quad $a=\sum_{n=1}^N\overline{y}^n\underline{f}^n$\\
& \quad $b=\sum_{n=1}^N\underline{f}^n$ & \quad $b=\sum_{n=1}^N\underline{f}^n$\\
& \quad $L=0$ & \quad $R=N$\\
2 & Compute & Compute \\
& \quad $L=L+1$ & \quad $a=a+\overline{y}^R(\overline{f}^R-\underline{f}^R)$\\
& \quad $a=a+\underline{y}^L(\overline{f}^L-\underline{f}^L)$ & \quad $b=b+\overline{f}^R-\underline{f}^R$ \\
& \quad $b=b+\overline{f}^L-\underline{f}^L$ & \quad $y_r=a/b$ \\
& \quad $y_l=a/b$ & \quad $R=R-1$ \\
3 & If $y_l\le y^{L+1}$, stop; & If $y_r\ge y^R$, stop; \\
 & otherwise, go to Step 2. &  otherwise, go to Step 2. \\ \hline
\end{tabular}
\end{table}

Once $y_l(\mathbf{x'})$ and $y_r(\mathbf{x'})$ are obtained, the final defuzzified output is:
\begin{align}
  y(\mathbf{x'}) = \frac{y_l(\mathbf{x'})+y_r(\mathbf{x'})} {2}. \label{eq:KMTR}
\end{align}

\subsubsection{Direct Defuzzification} \label{sect:def}

There are also many proposals to by-pass type-reduction\footnote{Although the type-reduced set provides a measure of the uncertainties that have flowed through all of the IT2 fuzzy system computations, it does not have to be (and almost never has been) used in practical applications.} and compute the defuzzified output directly \cite{Begian2008,Coupland2007b,Du2010,Gorzalczany1987b,Greenfield2008,Mendel2013,Nie2008,Tao2012}. A comprehensive description and comparison is also given in  \cite{drwuCCTFS2013}. The two most popular ones are the Nie-Tan (NT) method \cite{Begian2008,Nie2008}, which computes the output as
\begin{align}
y(\mathbf{x'})=\frac{\sum_{n=1}^N y^n[\underline{f}^n(\mathbf{x'})+\overline{f}^n(\mathbf{x'})]} {\sum_{n=1}^N[\underline{f}^n(\mathbf{x'})+\overline{f}^n(\mathbf{x'})]}, \label{eq:NT}
\end{align}
and the Begian-Melek-Mendel (BMM) method \cite{Begian2008}, which computes the output as
\begin{align}
y(\mathbf{x'})=\alpha\frac{\sum_{n=1}^N y^n\underline{f}^n(\mathbf{x'})}{\sum_{n=1}^N\underline{f}^n(\mathbf{x'})}+
\beta\frac{\sum_{n=1}^N y^n\overline{f}^n(\mathbf{x'})}{\sum_{n=1}^N\overline{f}^n(\mathbf{x'})}, \label{eq:BMM}
\end{align}
where $\alpha$ and $\beta$ are adjustable coefficients.

\subsection{Optimization: Gradient-Based Methods or Evolutionary Computation Methods}

Because both T1 and IT2 fuzzy systems have many parameters to optimize, as shown in Table~\ref{tab:P}, it is very difficult to tune them manually. Automatic optimization is usually needed.

\begin{table*}[htpb]\centering
\caption{The parameters and numbers of parameters in four different fuzzy systems (with $N$ rules and $p$ antecedents in each rule, i.e. $n = 1,\ldots, N$ and $k = 1,\ldots, p$). All T1 fuzzy sets are Gaussian (Fig.~\ref{fig:4}) and all IT2 fuzzy sets are Gaussian with uncertain means (Fig.~\ref{fig:4}) or uncertain standard deviations. The T1 Mamdani fuzzy system uses center-of-sets defuzzifier (Fig.~\ref{fig:7}), and the IT2 Mamdani fuzzy system uses center-of-sets type-reduction + Defuzzification (Fig.~\ref{fig:7}).} \setlength{\tabcolsep}{0.1cm}
\begin{tabular}{c|ccc} \hline
Fuzzy  & Parameters in  & Parameters in  & Total number  \\
system &  one antecedent & one consequent & of parameters \\ \hline
T1 Mamdani & $m_{X_k^n}$, $\sigma_{X_k^n}$ & $y^n$ & $(2p+1)N$ \\
IT2 Mamdani & $m_{k,1}^n$, $m_{k,2}^n$, $\sigma_k^n$; or  $m_{k}^n$, $\sigma_{k,1}^n$, $\sigma_{k,2}^n$ & $\underline{y}^n$, $\overline{y}^n$ & $(3p+3)N$ \\
T1 TSK & $m_{X_k^n}$, $\sigma_{X_k^n}$ & $\{c_k^n\}_{k=0}^p$ & $(3p+1)N$\\
IT2 TSK & $m_{k,1}^n$, $m_{k,2}^n$, $\sigma_k^s$; or  $m_{k}^n$, $\sigma_{k,1}^n$, $\sigma_{k,2}^n$ &$\{\underline{c}_k^n\}_{k=0}^p$, $\{\overline{c}_k^n\}_{k=0}^p$ & $(5p+2)N$\\ \hline
  \end{tabular}   \label{tab:P}
\end{table*}

Generally there are four major categories of optimization algorithms: exhaustive grid search, mathematical programming, gradient-based algorithms, and heuristic algorithms, particularly evolutionary computation (EC) algorithms. The optimization of a fuzzy system may use one of them, or a combination of more than one approaches. For example, the popular ANFIS model for T1 fuzzy systems \cite{Jang1993} combines mathematical programming (least squares estimation) and a gradient-based algorithm in its optimization. EC algorithms are recommended for the optimization of IT2 fuzzy systems, because derivatives are difficult to compute in an IT2 fuzzy system (especially when the LMF and/or UMF formulas include tests about the location of their independent variable), and such algorithms are globally convergent \cite{Mendel2014,Cervantes2018,Gonzalez2016,Sanchez2015a,Martinez-Soto2014}. There are many such EC algorithms, e.g., genetic algorithms, simulated annealing, particle swarm optimization, etc.

\begin{figure*}[htpb]\centering
\includegraphics[width=.8\linewidth]{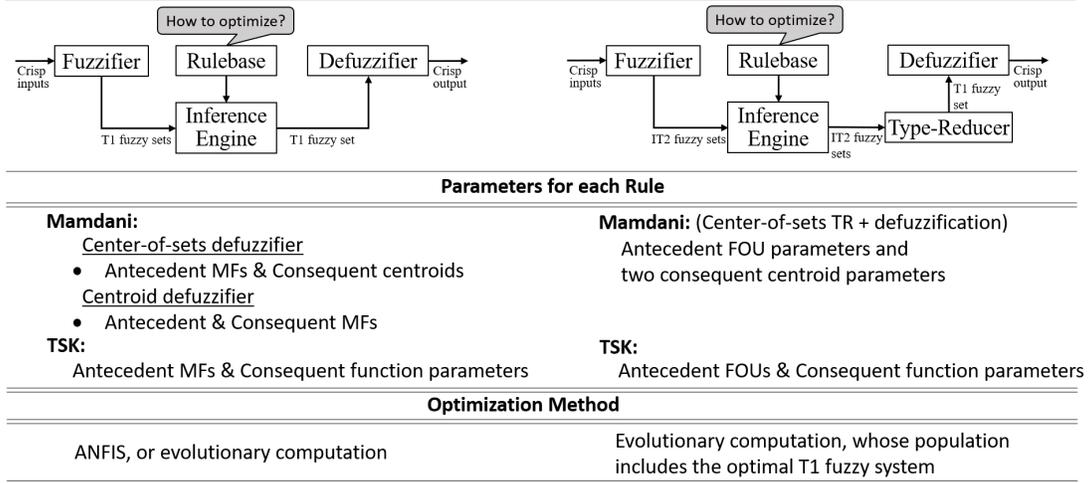} \caption{Optimization of fuzzy systems.} \label{fig:8}
\end{figure*}

Once an EC algorithm is chosen, there are two tuning strategies for an IT2 fuzzy system: \emph{one-step (totally independent) approach}, where an IT2 fuzzy system is tuned from scratch, or \emph{two-step (partially dependent) approach}, where an optimal baseline T1 fuzzy system is tuned first and then optimal FOUs are added to it. In the two-step approach one can include the optimal T1 fuzzy system in the population for the IT2 fuzzy system, which guarantees that the performance of the resulting IT2 fuzzy system is at least as good as the T1 fuzzy system. Details on how to do that using quantum particle swarm optimization are given in \cite{Mendel2014,Mendel2017}. Additionally, the two-step approach also reveals how much performance improvement an IT2 fuzzy system gets over the optimal T1 fuzzy system, and hence is very useful to practitioners: if the performance improvement is not significant, a practitioner may choose to use the T1 fuzzy system for simplicity and speed. For these reasons, we recommend the two-step approach for optimizing IT2 fuzzy systems, and EC algorithms that can include the optimal T1 fuzzy system in the population, e.g., genetic algorithms and particle swarm optimization (Fig.~\ref{fig:8}).

\subsection{Summary}

\begin{figure}[htpb]\centering
  \subfigure[]{\label{fig:9a} \includegraphics[width=.8\linewidth]{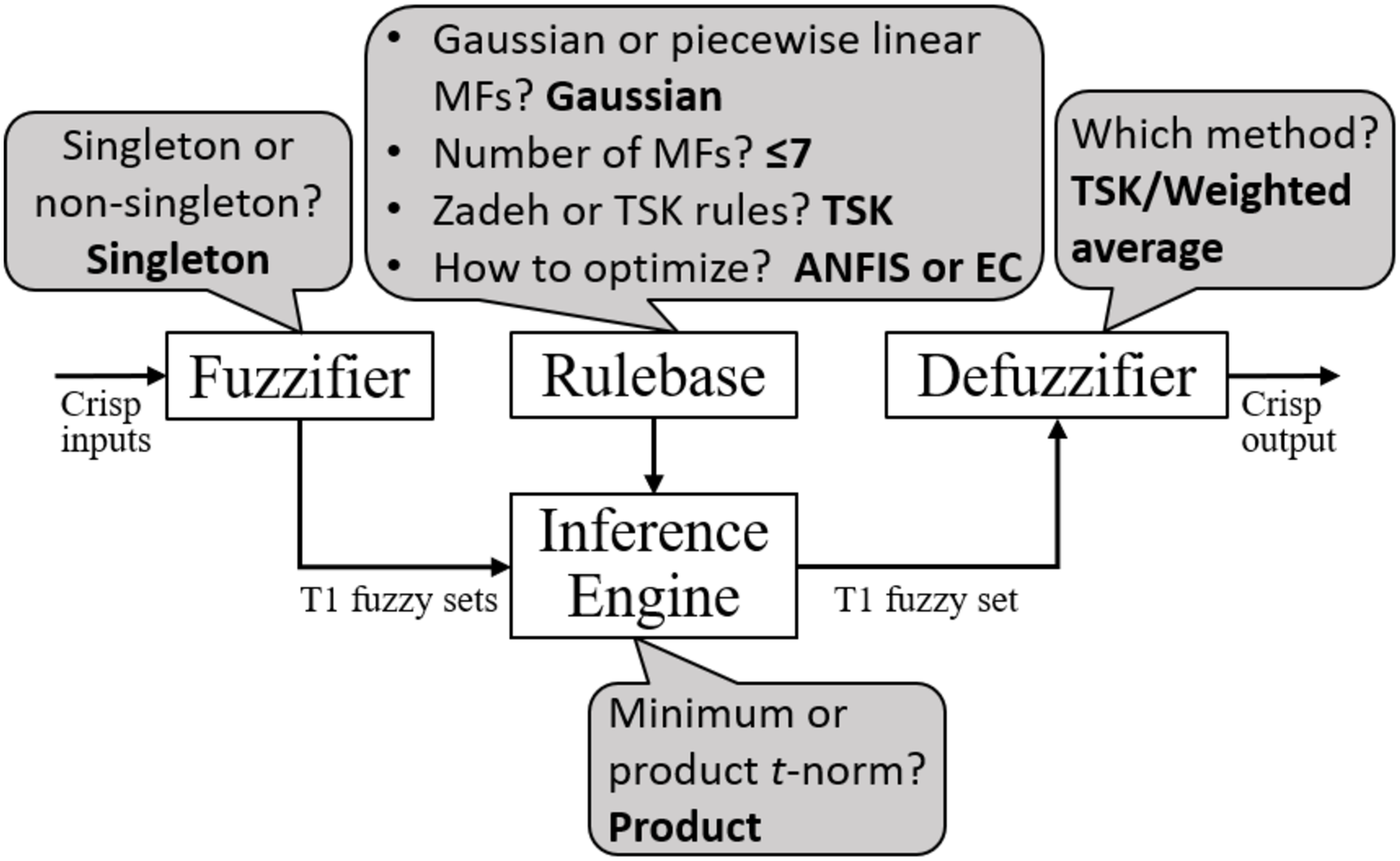}}
  \subfigure[]{\label{fig:9b} \includegraphics[width=.8\linewidth]{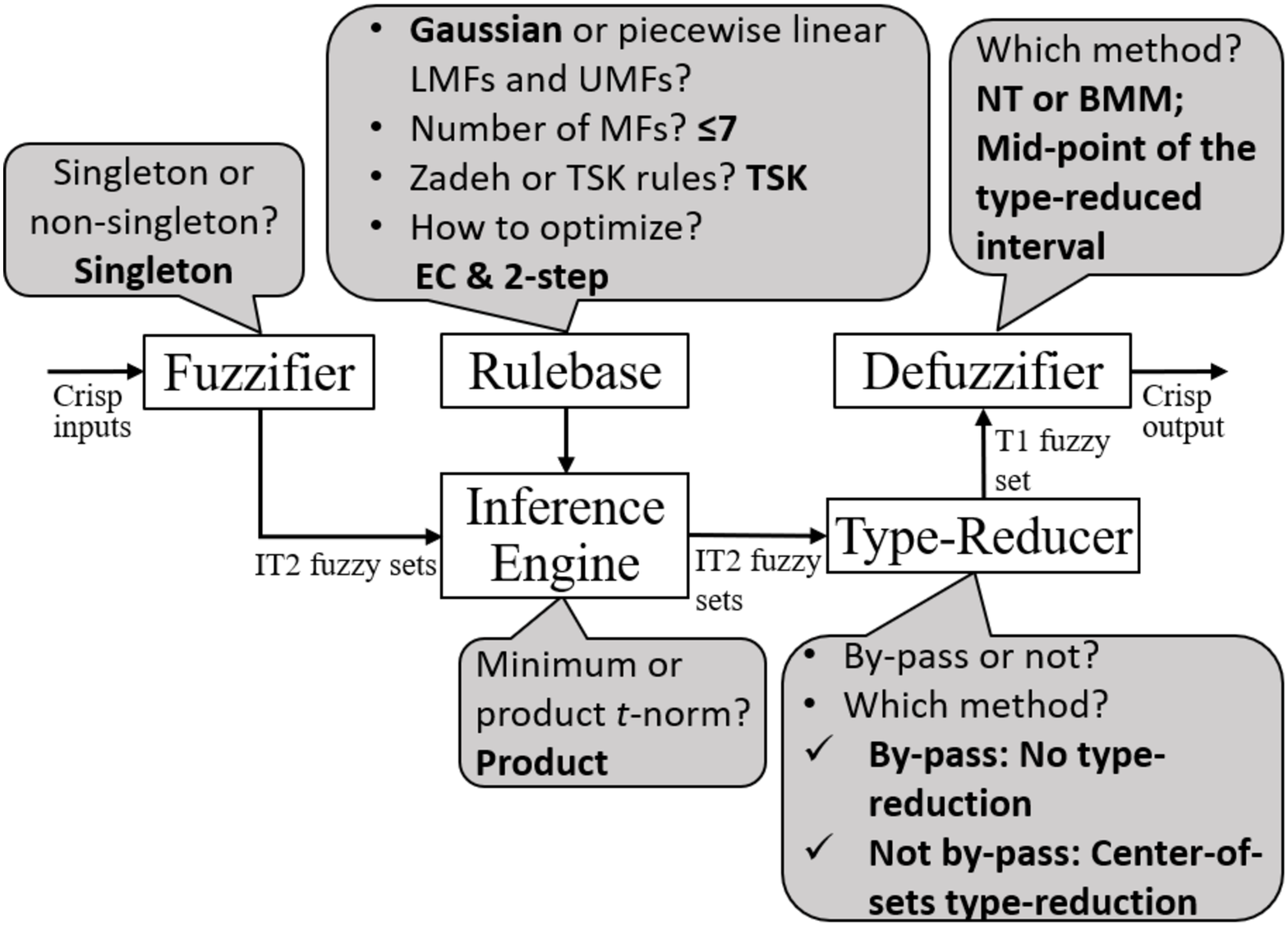}}
\caption{Summary of recommendations for designing (a) T1 and (b) IT2 fuzzy systems.} \label{fig:9}
\end{figure}

Our recommendations for practical T1 fuzzy system design are summarized in Fig.~\ref{fig:9a}: we recommend the simplest singleton TSK T1 fuzzy system (each rule consequent is a constant instead of a function of the inputs), with $\le7$ Gaussian MFs in each input domain, product $t$-norm, weighted average defuzzification, and ANFIS (when applicable, e.g., in function approximation applications) or EC (when the derivatives are difficult to compute, e.g., in fuzzy controller design) optimization.

Our recommendations for practical IT2 fuzzy system design are summarized in Fig.~\ref{fig:9b}: we recommend the simplest singleton TSK IT2 fuzzy system (each rule consequent is a constant), with $\le7$ Gaussian FOUs in each input domain, and the product $t$-norm. The output can be computed by using: 1) EIASC for center-of-sets type-reduction and then defuzzification; or, 2) the NT or BMM method directly. The optimization should be done by EC algorithms using a two-step approach.

More specifically, assume an IT2 fuzzy system has $p$ inputs and $N$ rules of the form (\ref{eq:Rule}). For an input vector $\mathbf{x}'=(x_1',x_2',...,x_p')$, our recommended procedure for computing the output of the IT2 fuzzy system is:
\begin{enumerate}
\item Compute the membership interval of $x_i'$ for each $\widetilde{X}_i^n$, $[\mu_{\underline{X}_i^n}(x_i'),\,\mu_{\overline{X}_i^n}(x_i')]$, $i=1,2,...,p$ and $n=1,2,...,N$.

\item Compute the firing interval of the $n^{\mathrm{th}}$ rule, $F^n$, using the product $t$-norm:
\begin{align}
  F^n(\mathbf{x}') &= [\mu_{\underline{X}_1^n}(x_1')\times\cdots\times\mu_{\underline{X}_I^n}(x_I'),\nonumber \\& \hspace*{6mm} \mu_{\overline{X}_1^n}(x_1')\times\cdots\times\mu_{\overline{X}_I^n}(x_I')]\nonumber \\
  & \equiv [\underline{f}^n,\,\overline{f}^n], \quad n=1,...,N  \label{eq2:fire_level}
\end{align}

\item Compute the output by combining $F^n(\mathbf{x}')$ and the corresponding rule consequents. This can be done by using EIASC in center-of-sets type-reduction and defuzzification separately, or by direct defuzzification using (\ref{eq:NT}) or (\ref{eq:BMM}).
\end{enumerate}

\subsection{Example}

The following proportional-integral (PI) fuzzy controller is used to illustrate the computations for a T1 fuzzy system and also an IT2 fuzzy system, following our above recommendations.

\begin{figure}[htpb]  \centering
  \subfigure[]{\label{fig:x1e1}     \includegraphics[width=.6\linewidth]{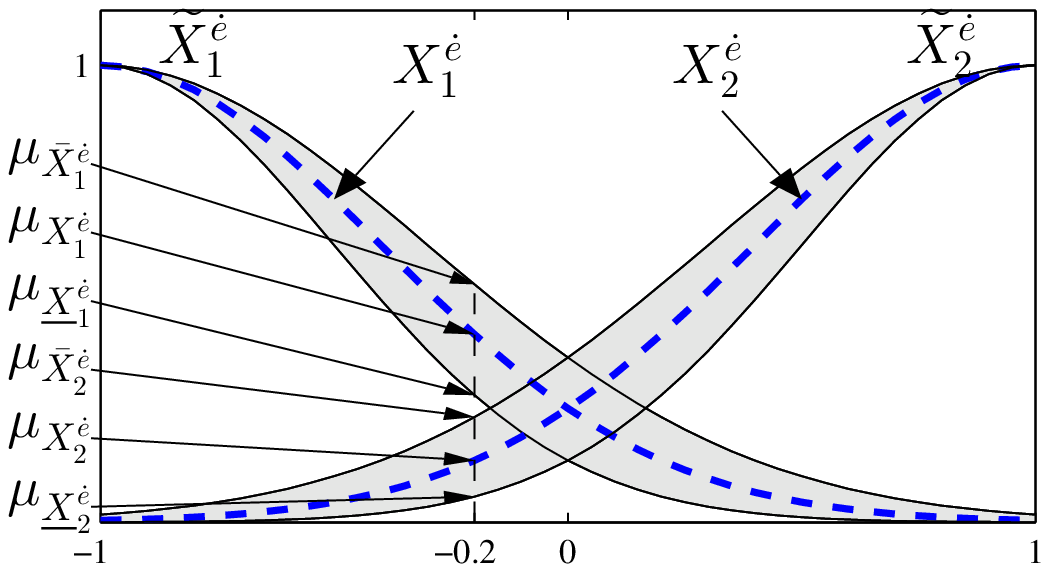}}
  \subfigure[]{\label{fig:x2e1}     \includegraphics[width=.6\linewidth]{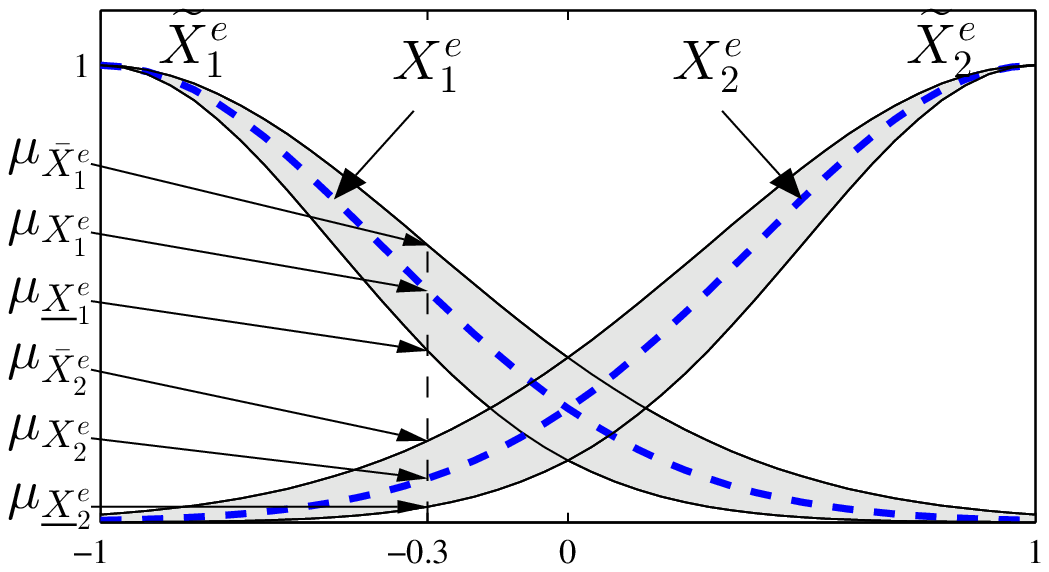}}
  \caption{Firing levels of the T1 fuzzy controller, and firing intervals of the IT2 fuzzy controller, when $\textbf{x}'=(\dot{e}',e')=(-0.2,-0.3)$. (a) MFs for $\dot{e}$, and (b) MFs for $e$.} \label{fig:e1}
\end{figure}

The MFs of the T1 PI fuzzy controller are shown in Fig.~\ref{fig:e1} as the bold dashed curves, where the centers of the Gaussian MFs are at $\pm1$, and all standard deviations are 0.6. Its four rules are:
\begin{align*}
   R^1: & \quad \mbox{IF } \dot{e} \mbox{ is } X_1^{\dot{e}} \mbox{ and } e \mbox{ is } X_1^e, \mbox{ THEN } \dot{u} \mbox{ is } y^1. \\
   R^2: & \quad\mbox{IF } \dot{e} \mbox{ is } X_1^{\dot{e}} \mbox{ and } e \mbox{ is } X_2^e, \mbox{ THEN } \dot{u} \mbox{ is } y^2. \\
   R^3: & \quad\mbox{IF } \dot{e} \mbox{ is } X_2^{\dot{e}} \mbox{ and } e \mbox{ is } X_1^e, \mbox{ THEN } \dot{u} \mbox{ is } y^3. \\
   R^4: & \quad\mbox{IF } \dot{e} \mbox{ is } X_2^{\dot{e}} \mbox{ and } e \mbox{ is } X_2^e, \mbox{ THEN } \dot{u} \mbox{ is } y^4.
 \end{align*}
where $\dot{u}$ is the \emph{change of the control signal}, $e$ is the \emph{feedback error}, and $\dot{e}$ is the \emph{change of error}. $y^1-y^4$ are given in Table~\ref{tab2:rulebase}.

\begin{table}[htpb]\centering
\caption{The rule consequents of the T1 and IT2 fuzzy controllers.} \setlength{\tabcolsep}{0.2cm} \renewcommand{\arraystretch}{1.3}
\begin{tabular}{c|cc} \hline\noalign{\smallskip}
 &  $X_1^e$ ($\widetilde{X}_1^e$)&  $X_2^e$ ($\widetilde{X}_2^e$) \\  \hline
        $X_1^{\dot{e}}$ ($\widetilde{X}_1^{\dot{e}}$) &  $y^1=-1$  &  $y^2=-0.5$  \\
        $X_2^{\dot{e}}$ ($\widetilde{X}_2^{\dot{e}}$) &  $y^3=.5$ & $y^4=1$\\ \noalign{\smallskip}\hline\noalign{\smallskip}
  \end{tabular}   \label{tab2:rulebase}
\end{table}

Consider an input vector $\textbf{x}'=(\dot{e}',\ e')=(-0.2,-0.3)$, as shown in Fig.~\ref{fig:e1}. The firing levels of the four T1 fuzzy sets are:
\begin{align*}
 \mu_{X_1^{\dot{e}}}(\dot{e}')&= \exp\left(-\frac{[-0.2-(-1)]^2}{2\times 0.6^2}\right)=0.4111\\   \mu_{X_2^{\dot{e}}}(\dot{e}')&= \exp\left(-\frac{(-0.2-1)^2}{2\times 0.6^2}\right)=0.1353\\
 \mu_{X_1^e}(e')&= \exp\left(-\frac{[-0.3-(-1)]^2}{2\times 0.6^2}\right)=0.5063\\
 \mu_{X_2^e}(e')&=\exp\left(-\frac{(-0.3-1)^2}{2\times 0.6^2}\right)=0.0956
\end{align*}
The firing levels of its four rules are shown in Table~\ref{tab:E21}. The output of the T1 fuzzy controller is
\begin{align*}
\dot{u}=\frac{f^1y^1+f^2y^2+f^3y^3+f^4y^4}{f^1+f^2+f^3+f^4}=-0.5491.
\end{align*}

\begin{table*}[htpb]
    \caption{Firing levels of the four rules of the T1 fuzzy controller.}   \renewcommand{\arraystretch}{1.4}   \centering
  \setlength{\tabcolsep}{0.1cm}
  \begin{tabular}{llll} \hline\noalign{\smallskip}
    Rule:& \multicolumn{1}{c}{Firing Level} &  & Rule  \\
     No.:&  &  & Consequent \\ \hline
    $R^1:$ & $f^1=\mu_{X_1^{\dot{e}}}(\dot{e}')\cdot
    \mu_{X_1^e}(e')=0.4111\times0.5063=0.2082$ & $\rightarrow$ & $y^1=-1$ \\
    $R^2:$ & $f^2=\mu_{X_1^{\dot{e}}}(\dot{e}')\cdot
    \mu_{X_2^e}(e')=0.4111\times 0.0956=0.0556 $ & $\rightarrow$ & $y^2=-0.5$ \\
    $R^3:$ & $f^3=\mu_{X_2^{\dot{e}}}(\dot{e}')\cdot
    \mu_{X_1^e}(e')=0.1353\times0.5063=0.0685$ & $\rightarrow$ & $y^3=0.5$ \\
    $R^4:$ & $f^4=\mu_{X_2^{\dot{e}}}(\dot{e}')\cdot
    \mu_{X_2^e}(e')=0.1353\times0.0956=0.0129$ & $\rightarrow$ & $y^4=1$ \\ \hline\noalign{\smallskip}
  \end{tabular} \label{tab:E21}
\end{table*}

An IT2 PI fuzzy controller may be constructed by blurring the T1 fuzzy sets of a T1 fuzzy controller to IT2 fuzzy sets. In this example we blur the standard deviation of the T1 Gaussian MFs from 0.6 to an interval $[0.5,\, 0.7]$, as shown in Fig.~\ref{fig:e1}. The rulebase of the IT2 fuzzy controller is
\begin{align*}
   \widetilde{R}^1: & \quad \mbox{IF } \dot{e} \mbox{ is } \widetilde{X}_1^{\dot{e}} \mbox{ and } e \mbox{ is } \widetilde{X}_1^e, \mbox{ THEN } \dot{u} \mbox{ is } y^1. \\
   \widetilde{R}^2: & \quad\mbox{IF } \dot{e} \mbox{ is } \widetilde{X}_1^{\dot{e}} \mbox{ and } e \mbox{ is } \widetilde{X}_2^e, \mbox{ THEN } \dot{u} \mbox{ is } y^2. \\
   \widetilde{R}^3: & \quad\mbox{IF } \dot{e} \mbox{ is } \widetilde{X}_2^{\dot{e}} \mbox{ and } e \mbox{ is } \widetilde{X}_1^e, \mbox{ THEN } \dot{u} \mbox{ is } y^3. \\
   \widetilde{R}^4: & \quad\mbox{IF } \dot{e} \mbox{ is } \widetilde{X}_2^{\dot{e}} \mbox{ and } e \mbox{ is } \widetilde{X}_2^e, \mbox{ THEN } \dot{u} \mbox{ is } y^4.
 \end{align*}
$y^1-y^4$ have been given in Table~\ref{tab2:rulebase}.

Consider again the input vector $\textbf{x}'=(\dot{e}',\ e')=(-0.2,-0.3)$, as shown in Fig.~\ref{fig:e1}. The firing intervals of the four IT2 fuzzy sets are:
\begin{align*}
 \left[\mu_{\underline{X}_1^{\dot{e}}}(\dot{e}'),\ \mu_{\overline{X}_1^{\dot{e}}}(\dot{e}')\right]&=
 \left[\exp\left(-\frac{[-0.2-(-1)]^2}{2\times 0.5^2}\right),\right.\\
 &\hspace*{6mm} \left. \exp\left(-\frac{[-0.2-(-1)]^2}{2\times 0.7^2}\right)\right]\\
 &= [0.2780,\,0.5205]\\
  \left[\mu_{\underline{X}_2^{\dot{e}}}(\dot{e}'),\ \mu_{\overline{X}_2^{\dot{e}}}(\dot{e}')\right]&=
  \left[\exp\left(-\frac{(-0.2-1)^2}{2\times 0.5^2}\right),\right.\\
 &\hspace*{6mm}  
 \left.\exp\left(-\frac{[(-0.2-1)^2}{2\times 0.7^2}\right)\right]\\
 &= [0.0561,\,0.2301] \\
 \left[\mu_{\underline{X}_1^e}(e'),\ \mu_{\overline{X}_1^e}(e')\right]&=
 \left[\exp\left(-\frac{[-0.3-(-1)]^2}{2\times 0.5^2}\right),\right.\\
 &\hspace*{6mm}  
 \left.\exp\left(-\frac{[-0.3-(-1)]^2}{2\times 0.7^2}\right)\right]\\
 &= [0.3753,\,0.6065]\\
 \left[\mu_{\underline{X}_2^e}(e'),\ \mu_{\overline{X}_2^e}(e')\right]&=
 \left[\exp\left(-\frac{(-0.3-1)^2}{2\times 0.5^2}\right),\right.\\
 &\hspace*{6mm} \left.\exp\left(-\frac{[(-0.3-1)^2}{2\times 0.7^2}\right)\right]\\
 &=[0.0340,\,0.1783]
\end{align*}

The firing intervals of the four rules are shown in Table~\ref{tab:E2}. When type-reduction and defuzzification are performed separately, the EIASC algorithm gives $y_l=-0.8846$ and $y_r=0.0058$, and the final defuzzified output is $\dot{u} =\frac{y_l+y_r}{2}=-0.4394$.

\begin{table*}[htpb]
    \caption{Firing intervals of the four rules of the IT2 fuzzy controller.}   \renewcommand{\arraystretch}{1.3}   \centering
  \setlength{\tabcolsep}{0.1cm}
  \begin{tabular}{llll} \hline\noalign{\smallskip}
    Rule & \multicolumn{1}{c}{Firing Interval} &  & Rule  \\
    No.:&  &  & Consequent \\  \hline
    $\widetilde{R}^1:$ & $[\underline{f}^1,\ \overline{f}^1]=[\mu_{\underline{X}_1^{\dot{e}}}(\dot{e}')\cdot
    \mu_{\underline{X}_1^e}(e'), \mu_{\overline{X}_1^{\dot{e}}}(\dot{e}')\cdot
    \mu_{\overline{X}_1^e}(e')]$ & $\rightarrow$ & $y^1=-1$ \\
    & $=[0.2780\times0.3753,0.5205\times0.6065]=[0.1044,\,0.3157]$ & &\\
    $\widetilde{R}^2:$ & $[\underline{f}^2,\ \overline{f}^2]=[\mu_{\underline{X}_1^{\dot{e}}}(\dot{e}')\cdot
    \mu_{\underline{X}_2^e}(e'), \mu_{\overline{X}_1^{\dot{e}}}(\dot{e}')\cdot
    \mu_{\overline{X}_2^e}(e')]$ & $\rightarrow$ & $y^2=-0.5$ \\
    &$=[0.2780\times0.0340,0.5205\times0.1783]=[0.0095,\,0.0928]$&&\\
    $\widetilde{R}^3:$ & $[\underline{f}^3,\ \overline{f}^3]=[\mu_{\underline{X}_2^{\dot{e}}}(\dot{e}')\cdot
    \mu_{\underline{X}_1^e}(e'), \mu_{\overline{X}_2^{\dot{e}}}(\dot{e}')\cdot
    \mu_{\overline{X}_1^e}(e')]$ & $\rightarrow$ & $y^3=0.5$ \\
    &$=[0.0561\times0.3753,0.2301\times0.6065]=[0.0211,\,0.1395]$&&\\
    $\widetilde{R}^4:$ & $[\underline{f}^4,\ \overline{f}^4]=[\mu_{\underline{X}_2^{\dot{e}}}(\dot{e}')\cdot
    \mu_{\underline{X}_2^e}(e'), \mu_{\overline{X}_2^{\dot{e}}}(\dot{e}')\cdot
    \mu_{\overline{X}_2^e}(e')]$ & $\rightarrow$ & $y^4=1$ \\
    & $=[0.0561\times0.0340,0.2301\times0.1783]=[0.0019,\,0.0410]$ & & \\ \hline\noalign{\smallskip}
  \end{tabular} \label{tab:E2}
\end{table*}

When (\ref{eq:NT}) is used, the final output is computed in (\ref{eq:dotu}). When (\ref{eq:BMM}) is used and $\alpha=\beta=0.5$, the final output is computed in (\ref{eq:dotu2}).

\begin{figure*}
\begin{align}
\dot{u}&=\frac{-(0.1044+0.3157)-0.5(0.0095+0.0928)+0.5(0.0211+0.1395)+(0.0019+0.0410)}
{(0.1044+0.3157)+(0.0095+0.0928)+(0.0211+0.1395)+(0.0019+0.0410)}=-0.4794 \label{eq:dotu}\\
\dot{u}&=0.5\times \frac{-1\times0.1044-0.5\times0.0095+0.5\times0.0211+1\times0.0019}
{0.1044+0.0095+0.0211+0.0019}\nonumber \\
&\quad +0.5\times \frac{-1\times0.3157-0.5\times0.0928+0.5\times0.1395+1\times0.0410}
{0.3157+0.0928+0.1395+0.0410}=-0.5665 \label{eq:dotu2}
\end{align}
\end{figure*}

Note that the outputs computed from the three approaches are different, given the same rules and MFs. However, in practice one first decides which method to use, and then tunes the rules and MFs accordingly so as to optimize a performance matric (e.g., root mean squared error in function approximation, or time-weighted integral absolute error in controls \cite{drwuEAAI2006,drwuISA2006}). In general, the optimal rules and MFs for the three methods are different.

More examples on the detailed computations of T1 and IT2 fuzzy systems, including many that are not introduced in this paper (due to page limit and targeted readers), can be found in \cite{Mendel2017}.

\subsection{Software}

Matlab has a Fuzzy Logic Toolbox which considers only T1 fuzzy systems. Several researchers have developed their own Matlab toolboxes/packages for IT2 fuzzy systems, e.g., Mendel's software\footnote{http://sipi.usc.edu/$\sim$mendel}, Wu's functions\footnote{https://www.mathworks.com/matlabcentral/fileexchange/29006-functions-for-interval-type-2-fuzzy-logic-systems}, and Taskin and Kumbasar's open source Matlab/Simulink Toolbox\footnote{http://web.itu.edu.tr/kumbasart/type2fuzzy.htm} \cite{Taskin2015}. Additionally, Wagner developed a Java based toolkit, Juzzy\footnote{http://juzzy.wagnerweb.net/}, for T1, IT2 and general T2 fuzzy systems.

A beginner can start with Wu's functions because they concisely illustrate the essentials of IT2 fuzzy systems by a simple example.

\section{Myths about IT2 Fuzzy Systems}

The successful applications of IT2 fuzzy systems have created some myths about their performance. Here we shall clarify two of them. To do this we will use IT2 fuzzy controllers as an example in the illustrations, but the clarifications can also be extended to other applications of IT2 fuzzy systems.

\subsection{Myth 1: Changing T1 Fuzzy Sets to IT2 Fuzzy Sets Automatically Improves Performance}

Many applications have shown that IT2 fuzzy controllers can achieve better control performance than their T1 counterparts. This has been attributed to the FOUs, which lead to two fundamental differences between T1 and IT2 fuzzy controllers \cite{drwuFundamental2012}: 1) \emph{Adaptiveness}, meaning that the embedded T1 fuzzy sets used to compute the bounds of the type-reduced interval change as input changes; and, 2) \emph{Novelty}, meaning that the upper and lower MFs of the same IT2 fuzzy set may be used simultaneously in computing each bound of the type-reduced interval. As a result, an IT2 fuzzy controller beginner may get the impression that by changing T1 fuzzy sets to IT2 fuzzy sets the resulting IT2 fuzzy controller will automatically have better performance. Unfortunately, this is not the case. Carefully designed FOUs may improve performance [see \cite{Mendel2018} for further discussions), but arbitrary FOUs almost never do. To achieve better performance, one needs to re-tune the IT2 fuzzy controller, either from scratch, or using the T1 fuzzy controller as a baseline \cite{Mendel2014,drwuEAAI2006}. There is no black magic that, by changing T1 fuzzy sets to IT2 fuzzy sets, an IT2 fuzzy controller will automatically outperform a T1 one.


\subsection{Myth 2: Optimizing an IT2 Fuzzy Controller in Known Scenarios Guarantees Its Optimal Performance in an Unknown Scenario}

We have seen many cases where people tune an IT2 fuzzy controller for some operating conditions but then apply it to different operating conditions, and claim that its performance is not as good as expected. This is because the design procedure is not correct. If one wants the IT2 fuzzy controller to have good performance under a variety of operating conditions, then all these conditions must be considered during the tuning phase. This is analogous to the well-known fact in machine learning: when a machine learning model is trained on a specific dataset, but tested on a different dataset, it is not likely to perform well. In machine learning, we require the training dataset to be as complete and diverse as possible for good test performance. In IT2 fuzzy controller design, we also need to ensure the scenarios considered in training are complete and diverse enough.

For example, in \cite{drwuEAAI2006} Wu and Tan wanted the IT2 fuzzy controller to be able to respond quickly to setpoint changes, and also to robustly handle modeling uncertainties including time delay and parameter variations of the underlying physical model. All these different scenarios were considered during the design phase. As a result, experimental results were consistent with simulation results. If any of those scenarios was not considered in the design phase, e.g., if the IT2 fuzzy controller was tuned without considering time delay, but was then applied to a plant with time delay, then very likely the performance would have been much worse.

\section{Conclusions}

There are many choices to be made in designing a well-performing IT2 fuzzy system, including the kind of fuzzifier (singleton or non-singleton), number of membership functions (MFs), shape of MFs (Gaussian or piecewise linear), kind of rules (Zadeh or TSK), kind of inference (minimum or product $t$-norm), kind of output processing (type-reduction or not), and method for optimization. While these choices give an experienced IT2 fuzzy system researcher extensive freedom to design the optimal IT2 fuzzy system, they may look overwhelming and confusing to IT2 beginners.

To help the IT2 beginner overcome the learning barrier, this paper recommends the following representative starting choices for IT2 fuzzy systems based on our experience: singleton fuzzifier, simplest TSK rules, $\le$ 7 Gaussian FOUs in each input domain, product $t$-norm, computing the output by using the EIASC algorithms for center-of-sets type-reduction and then defuzzification, or, the NT or BMM method directly, and a two-step EC algorithm for optimization. We are not claiming our recommendations are always the best. An experienced researcher on IT2 fuzzy systems may be able to design better IT2 fuzzy systems by using other choices; however, our recommendations have a high chance of leading to IT2 fuzzy systems that can outperform T1 ones. We have also clarified two myths about IT2 fuzzy systems.

We hope that this paper will be very useful to IT2 beginners, and will also help to promote wider applications of IT2 fuzzy systems.

\section*{Acknowledgement}
This research was supported by the National Natural Science Foundation of China Grant 61873321, and the 111 Project on Computational Intelligence and Intelligent Control under Grant B18024.

\section*{References}

\end{document}